\documentclass[journal]{IEEEtran}
\usepackage{color}
\usepackage{cite}
\usepackage{ifpdf}
\usepackage{array}
\usepackage{tabularray}
\usepackage{multirow}
\usepackage{relsize}
\usepackage[table,xcdraw]{xcolor}
\usepackage{url}
\ifCLASSINFOpdf
  \usepackage[pdftex]{graphicx}
\else
  \usepackage[dvips]{graphicx}
\fi

\usepackage{amsmath}
\usepackage{algorithmic}

\begin{document}

\title{RadCLIP: Enhancing Radiologic Image Analysis through
Contrastive Language-Image Pre-training}

\author{%
  Zhixiu Lu\textsuperscript{1}\thanks{Submitting author.}\,,
  Hailong Li\textsuperscript{1,2,3,4}\,,
  Nehal A. Parikh\textsuperscript{3,5}\,,
  Jonathan R. Dillman\textsuperscript{1,2,4}\,,
  Lili He\textsuperscript{1,2,3,4,5,6,7,8}\thanks{Corresponding author: \texttt{lili.he@cchmc.org}.}%
  \\[1ex]
  \textsuperscript{1}Imaging Research Center, Department of Radiology, Cincinnati Children’s Hospital Medical Center, Cincinnati, OH, USA\\
  \textsuperscript{2}Artificial Intelligence Imaging Research Center, Cincinnati Children’s Hospital Medical Center, Cincinnati, OH, USA\\
  \textsuperscript{3}Neurodevelopmental Disorders Prevention Center, Perinatal Institute, Cincinnati Children’s Hospital Medical Center, Cincinnati, OH, USA\\
  \textsuperscript{4}Department of Radiology, University of Cincinnati College of Medicine, Cincinnati, OH, USA\\
  \textsuperscript{5}Department of Pediatrics, University of Cincinnati College of Medicine, Cincinnati, OH, USA\\
  \textsuperscript{6}Department of Computer Science, University of Cincinnati, Cincinnati, OH, USA\\
  \textsuperscript{7}Department of Biomedical Engineering, University of Cincinnati, Cincinnati, OH, USA\\
  \textsuperscript{8}Department of Biomedical Informatics, University of Cincinnati College of Medicine, Cincinnati, OH, USA\\[1ex]

}

\maketitle

\begin{abstract}

The integration of artificial intelligence (AI) with radiology signifies a transformative era in medicine. Vision foundation models have been adopted to enhance radiologic imaging analysis. However, the inherent complexities of 2D and 3D radiologic data present unique challenges that existing models, which are typically pre-trained on general non-medical images, do not adequately address. To bridge this gap and harness the diagnostic precision required in radiologic imaging, we introduce Radiologic Contrastive Language-Image Pre-training (RadCLIP): a crossmodal vision-language foundational model that utilizes a Vision Language Pre-training (VLP) framework to improve radiologic image analysis.

Building on the Contrastive Language-Image Pre-training (CLIP) approach, RadCLIP incorporates a slice pooling mechanism designed for volumetric image analysis and is pre-trained using a large, diverse dataset of radiologic image-text pairs. This pre-training effectively aligns radiologic images with their corresponding text annotations, resulting in a robust vision backbone for radiologic imaging. Extensive experiments demonstrate RadCLIP’s superior performance in both unimodal radiologic image classification and crossmodal image-text matching, underscoring its significant promise for enhancing diagnostic accuracy and efficiency in clinical settings.

Our key contributions include curating a large dataset featuring diverse radiologic 2D/3D image-text pairs, pre-training RadCLIP as a vision-language foundation model on this dataset, developing a slice pooling adapter with an attention mechanism for integrating 2D images, and conducting comprehensive evaluations of RadCLIP on various radiologic downstream tasks.
\end{abstract}

\begin{IEEEkeywords}
RadCLIP, Radiology, Foundation Model, Vision-Language Pretraining (VLP), Contrastive Language-Image Pre-training (CLIP), Medical Imaging, Representation Learning
\end{IEEEkeywords}

\IEEEpeerreviewmaketitle
\section{Introduction}
In the rapidly evolving field of radiology, integrating artificial intelligence (AI) has become indispensable. Vision foundation models trained on large datasets have shown promise in computer vision applications \cite{clip}. These models are a cornerstone for specialized applications. Transfer learning, where knowledge from one domain enhances performance in another, is particularly beneficial. In medical imaging, transfer learning is especially important given the difficulty of acquiring large radiologic datasets to train end-to-end deep learning models from scratch \cite{kim2022transfer}\cite{smith2022what}.

State-of-the-art vision foundation models are typically trained on natural image datasets such as CIFAR-10, Food-101, and ImageNet \cite{imagenet}. However, the unique challenges of radiologic imaging—its 2D/3D nature, subtle pathological features, and the high stakes of diagnostic errors—demand models tailored to the medical domain. Generic vision models trained on natural image datasets often fail to capture radiologic image intricacies, resulting in performance gaps. \cite{radimagenet}. For example, GPT-4V, one of the most prominent generic vision-language models, does not perform well on medical tasks \cite{Liu_Jiang_Zhong_Wu_Ma_Li_Yu_Zhang_Pan_Shu_et}.

Recent developments in vision-language models, which understand both images and text, have significantly improved the ability to associate images with words. Contrastive Language-Image Pre-training (CLIP), a pioneering work by OpenAI, leverages extensive image-text datasets for effective visual-textual concept association, enabling diverse applications such as zero-shot image recognition and advanced natural language tasks. Its adaptability and robustness underscore its foundational role. Alongside CLIP, models like CoCa and ALIGN have pushed the boundaries in crossmodal tasks and set new benchmarks, showcasing the potential of the vision-language pretraining (VLP) framework to enhance vision models through language supervision \cite{clip,coca,ALIGN}.

This crossmodal advancement has spurred innovations in areas such as video-text recognition \cite{wang-2023,rasheed2023finetuned}, crossmodal retrieval \cite{fang2021learningalignedcrossmodalrepresentation,Zhen_Hu_Peng_Goh_Zhou_2022a}, and visual question answering \cite{blip,pubmedclip}. Recently, efforts to adapt vision-language models for the medical domain \cite{zhang2024potentialmultimodallargelanguage} have led to several noteworthy projects. For example, CONVIRT automates radiology report generation using natural language processing, streamlining diagnostic workflows. GLoRIA leverages radiology reports to enhance image analysis without extensive labeling, using attention mechanisms to improve image retrieval, classification, and segmentation. MedCLIP adapts the CLIP framework to link chest X-rays (CXRs) with clinical notes, thereby boosting diagnostic accuracy in zero-shot learning. PubMedCLIP extracts information from medical literature to support clinical applications. CLIP-Lung integrates clinical text annotations with lung images to better predict 3D CT lung nodule malignancy via channel-wise condition prompting—one of the few methods extending VLP to 3D radiologic data. Finally, CXR-CLIP addresses CXR data scarcity by merging image-text and image-label data to learn study-level features using novel contrastive losses \cite{cxrclip,gloria,medclip,pubmedclip}.

These medicine-related vision-language models demonstrate the potential of the VLP framework in radiology. However, a major limitation is the lack of extensive, diverse radiologic imaging data for training and validation. Most existing models are developed using 2D CXRs or CT slices \cite{Moor2023}, which may limit their ability to capture heterogeneous imaging modalities. Typically, these datasets lack sufficient 3D data (e.g., 3D CT and MRI), a key attribute of radiologic imaging compared to natural image datasets like ImageNet \cite{imagenet}. This restricts their comprehensive understanding of 3D spatial information crucial for accurate diagnosis and assessment.

To address these limitations, we present Radiologic CLIP (RadCLIP), a novel vision-language model tailored for radiologic image analysis. RadCLIP overcomes current challenges by focusing on improved radiologic image representation learning. By leveraging a diverse, carefully curated 2D/3D radiologic dataset, we build a robust visual backbone and enhance crossmodal capabilities using the VLP framework. We evaluate RadCLIP on both unimodal image representation and crossmodal vision-language alignment tasks.

In summary, our contributions are:
\begin{enumerate}
  \item We collected and curated a large, diverse radiologic image-text dataset covering a wide range of 2D/3D modalities, anatomical regions, diseases, and conditions.
  \item We trained RadCLIP using these image-text pairs within the VLP framework.
  \item We introduced a slice-wise pooling mechanism for 3D images to integrate 2D slices, enhancing the model’s understanding of 3D spatial information.
  \item We conducted extensive experiments to evaluate RadCLIP’s performance in unimodal representation learning and crossmodal vision-language alignment.
\end{enumerate}
\section{Related Work}

\subsection{Radiologic Vision Foundation Models}

In recent years, the intersection of AI and radiology has garnered significant attention, spurring the development of models to enhance medical image analysis \cite{Srivastav2023, azad2023foundationalmodelsmedicalimaging}. Radiologic vision foundation models, trained on large radiologic datasets to capture diverse features, have shown exceptional promise in radiology tasks. One example is MedViT, a Vision Transformer for generalized medical image classification developed by Manzari et al \cite{medvit}. MedViT combines the strengths of Convolutional Neural Networks (CNNs) and Vision Transformers, addressing the quadratic complexity of self-attention while enhancing robustness against adversarial attacks by focusing on global structural features rather than textures. It also employs innovative data augmentation techniques that blend feature normalization with augmentation, resulting in superior accuracy across various medical imaging datasets.

Another significant development is RadImageNet and its associated foundation models \cite{radimagenet}. RadImageNet is a large-scale, domain-specific dataset comprising 1.35 million annotated CT, MRI, and Ultrasound images covering a broad range of pathological conditions. Studies have demonstrated that models pre-trained on RadImageNet outperform those trained on ImageNet for many medical imaging tasks, especially when data is scarce. For instance, RadImageNet models show marked improvements in analyzing thyroid nodules, breast masses, anterior cruciate ligament injuries, and meniscal tears, underscoring the importance of domain-specific datasets in enhancing AI performance in radiologic imaging.

\subsection{Radiologic Vision-Language Models}

Early adaptations of CLIP-like models to radiologic imaging have shown promise despite the challenges posed by the complexity of medical images and the nuanced language of radiology reports. Recent developments in this area include several radiologic vision-language models. For example, GLoRIA leverages radiology reports to learn detailed image representations without extensive manual labeling, significantly advancing label-efficient medical imaging \cite{gloria}. CONVIRT employs natural language processing to generate radiology reports that mimic expert annotations \cite{zhang2023adapting}. CXR-CLIP combines CXR-text and CXR-label data through class-specific prompts and introduces novel contrastive losses to capture study-level features \cite{cxrclip}. MedCLIP adapts the CLIP framework for CXRs by linking images with clinical notes, thereby enhancing zero-shot diagnostic accuracy \cite{medclip}. PMC-CLIP is designed to extract and correlate information from extensive medical literature, bridging the gap between academic research and clinical applications \cite{pmcclip}.

Despite these advances, a common limitation persists: most radiologic vision-language models are developed using 2D CXRs or CT slices, lacking the extensive, diverse 3D imaging data necessary to fully capture the heterogeneous modalities and spatial complexity of human anatomy. RadCLIP aims to address this gap by incorporating more diverse and comprehensive radiologic imaging data into its training and evaluation.
\section{Methodology}

\subsection{RadCLIP Vision-Language Pre-training}

CLIP has revolutionized the integration of vision and language by leveraging large-scale image-text datasets to learn rich, multimodal representations. Its dual-encoder architecture aligns visual and textual information in a shared embedding space by minimizing a contrastive loss that brings matching pairs closer while pushing apart mismatched pairs \cite{clip,virtexl}. Prior work shows that this VLP framework enables vision models to capture fine-grained image details through text supervision \cite{clip,blip,coca,flamingo}. Inspired by this success, RadCLIP employs the VLP framework to build a robust vision foundation model for radiologic image analysis via paired text supervision.

We train RadCLIP on a meticulously curated collection of radiologic image-text pairs covering a wide range of imaging modalities, anatomical regions, diseases, and conditions, ensuring robust and generalized performance. In addition, we introduce a novel slice pooling adapter with a slice-wise attention mechanism that weights individual image slices, thereby enhancing volumetric image analysis \cite{vaswani2017attention}. This module not only enables training a universal volumetric radiologic image encoder but also prioritizes the most informative slices.

\begin{figure*}[h!]
    \centering
    \includegraphics[height=6.8cm]{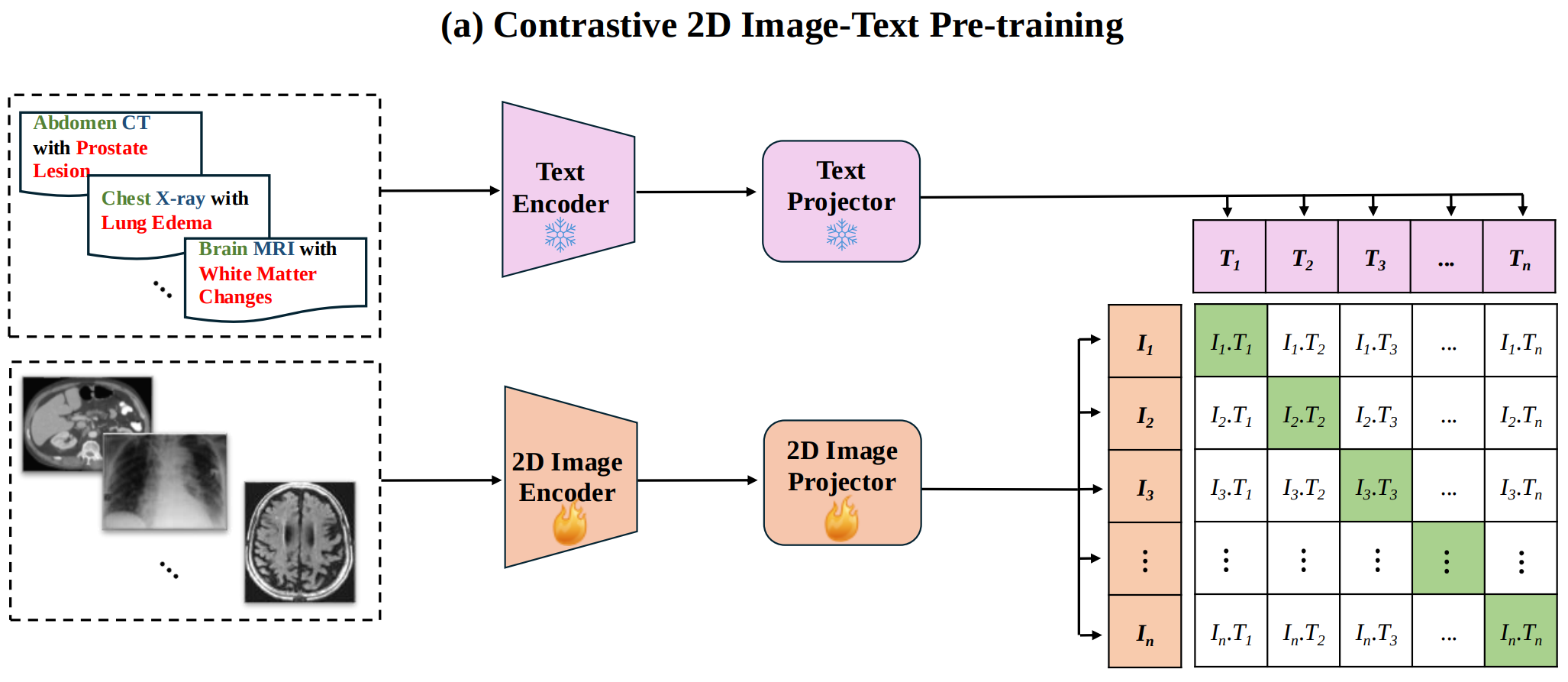} 
    \vspace{1em} 
    \includegraphics[height=6.8cm]{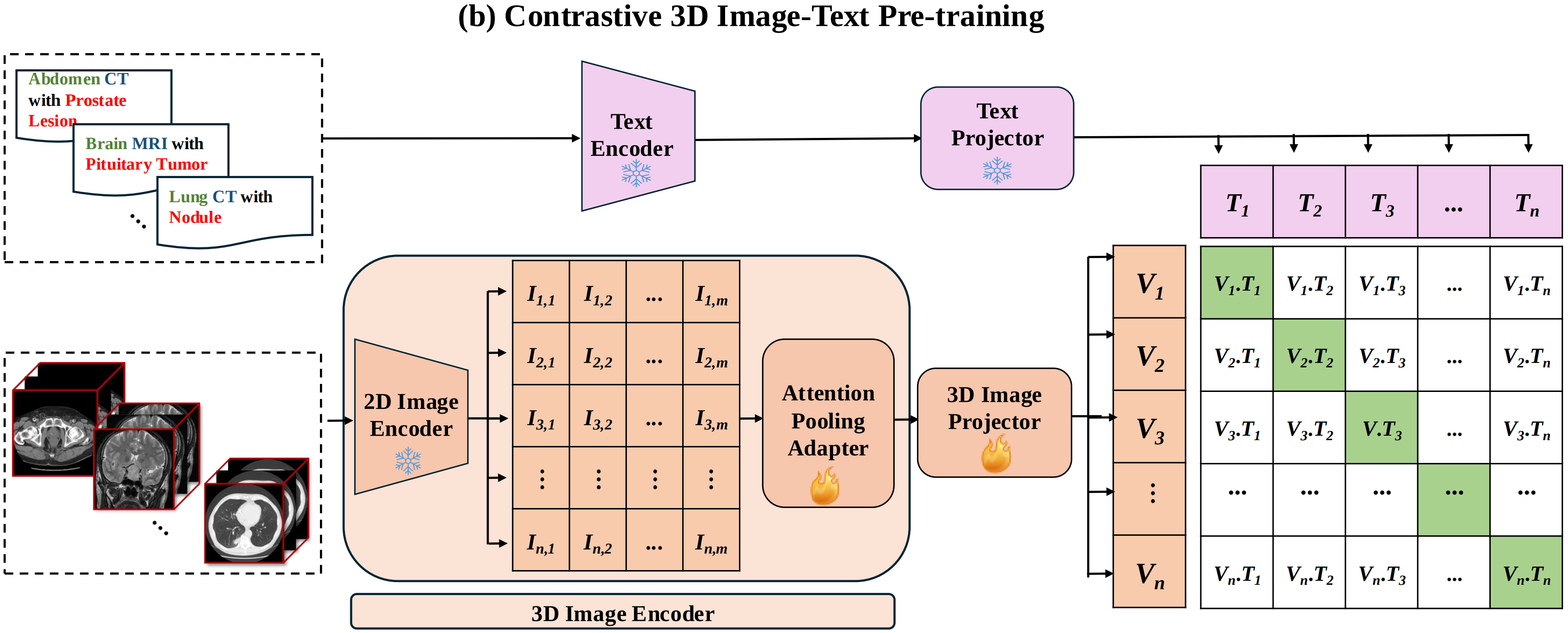} 
    \caption{RadCLIP Model Architecture. (a) The framework integrates a frozen text encoder from CLIP with a fine-tuned 2D image encoder to extract rich radiologic features. (b) The slice pooling adapter then aggregates these 2D slice embeddings into a unified 3D volumetric representation using an attention mechanism that preserves spatial context. Together, these components enable effective crossmodal alignment between radiologic images and their corresponding text descriptions.}
    \label{fig:model}
\end{figure*}

The RadCLIP architecture comprises three modules: a text encoder to process descriptions, a 2D image encoder for radiologic images, and a slice pooling adapter that aggregates 2D slice embeddings (see Figure \ref{fig:model}). We leverage the pre-trained CLIP text encoder and freeze its weights \cite{textgrounded}, while fine-tuning the 2D image encoder (Figure \ref{fig:model}a) and training the slice pooling adapter (Figure \ref{fig:model}b).

\subsubsection{2D Image Encoder Pre-training}

We fine-tune the pre-trained 2D image encoder from CLIP using contrastive pre-training on our large set of 2D radiologic image-text pairs. The encoder is trained to pull the embeddings of radiologic images \(\mathbf{I_i}\) and their corresponding text descriptions \(\mathbf{T_i}\) closer in the embedding space, while pushing apart mismatched pairs \(\mathbf{T_j}\). For example, an abdominal CT slice is pulled toward the text “Abdomen CT with Prostate Lesion” and pushed away from “Brain MRI with White Matter Changes.” The text encoder remains frozen to preserve its language understanding. This process enables the image encoder to learn meaningful representations for enhanced image-text alignment \cite{tsimpoukelli2021multimodal,Multi-Modality}.

\begin{figure}[ht]
\centering
\includegraphics[width=1.8in]{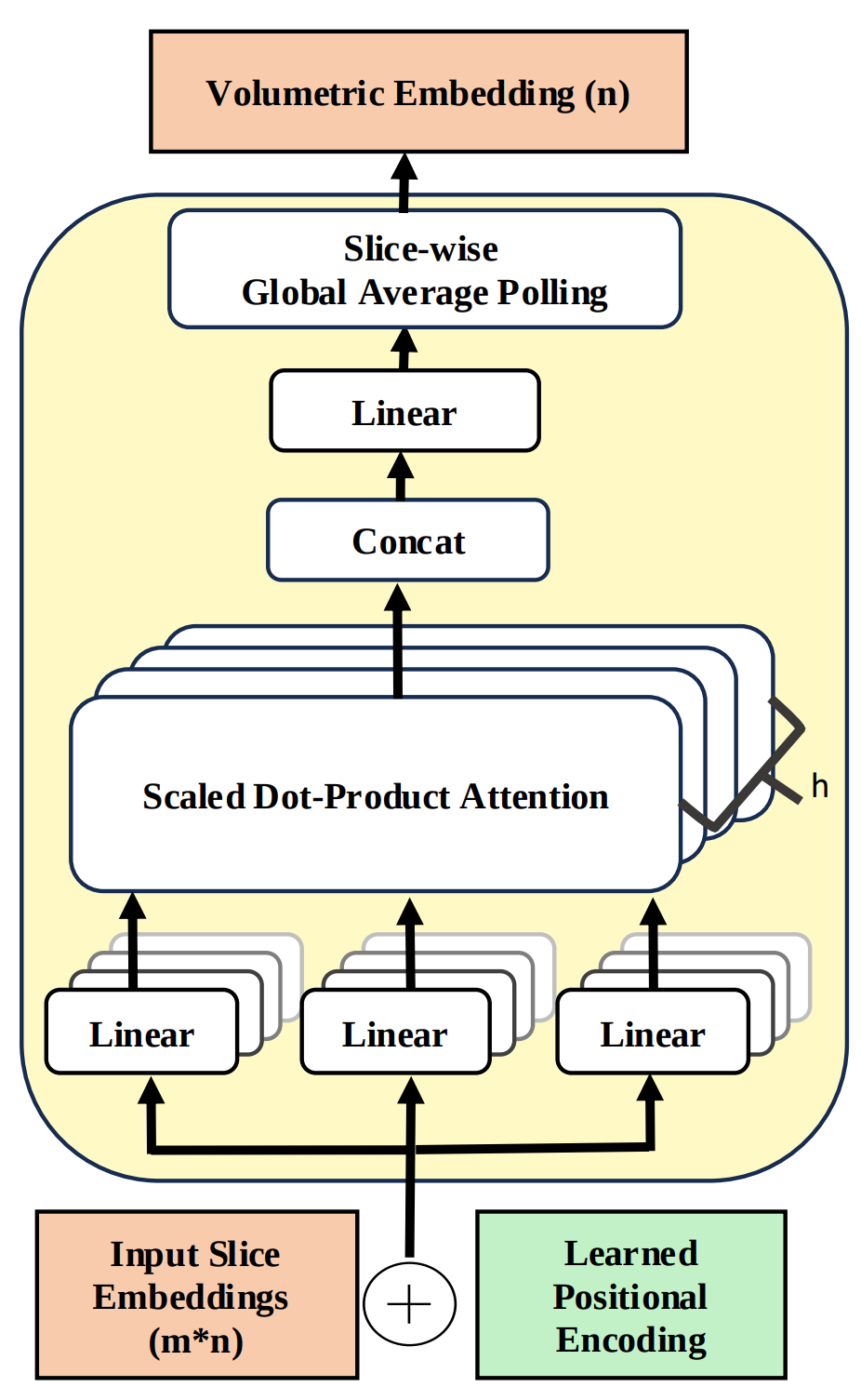}
\caption{This diagram details our adapter that converts a stack of 2D slice embeddings into a unified 3D image representation. The adapter employs a multi-head self-attention mechanism to capture inter-slice dependencies and integrates learnable random positional encoding to embed spatial order. }
\label{fig_SlicePooling}
\end{figure}

\subsubsection{Slice Pooling Adapter Pre-training}

For 3D volumetric radiologic images, traditional methods often use multi-channel feature maps or average pooling to aggregate 2D slice representations into a 3D volume \cite{shen2017deep,Vit2d3d,rasheed2023finetunedclipmodelsefficient}. However, such strategies can lead to information loss and insufficient context to capture complex anatomical structures \cite{LITJENS201760}. Recent research has explored more advanced adapter mechanisms, including 2D/3D convolutions \cite{liu-2023}, LSTMs \cite{wang-2023}, and attention-based pooling \cite{li-2023,Jun_Jeong_Heo_Suk_2024}.

Our approach introduces a slice pooling adapter that employs an attention-based pooling mechanism to integrate 2D slice representations into a unified 3D volume \cite{Wang_2019}. As shown in Figure \ref{fig_SlicePooling}, the adapter consists of a multi-head self-attention layer with learnable random positional encoding (PE) \cite{liu2020learningencodepositiontransformer}. This design overcomes the limitations of global average pooling by capturing critical spatial context while keeping the parameter count low.

Assuming \(\mathbf{I}\) represents a stack of 2D slice embeddings \(\mathbf{I_i}\) for \(i \in \{1,2,\ldots,n\}\), the volumetric representation is computed as:
\[
\mathbf{V} = \text{MHSA}(\mathbf{I} + \text{PE}(\mathbf{P}))
\]
where MHSA denotes the multi-head self-attention mechanism and \(\text{PE}(\mathbf{P})\) is the learnable random positional encoding applied to the slice indices. The encoding is defined as:
\[
\text{PE}(pos) = \text{LearnableRandom}(d_{\text{model}})
\]
with \(pos\) representing the positional index and \(d_{\text{model}}\) the model’s dimensionality.

The attention mechanism captures inter-slice relationships, providing a comprehensive understanding of volumetric data. Meanwhile, the learnable positional encoding (PE) facilitates adaptive learning of spatial positions, enhancing volume representation by integrating spatial information more effectively. We pre-trained the slice pooling adapter using contrastive learning on a diverse set of 3D radiologic image-text pairs. The adapter is trained to pull the embeddings of 3D volumetric images \(\mathbf{V_i}\) and their corresponding texts \(\mathbf{T_i}\) closer, while pushing apart mismatched pairs \(\mathbf{T_j}\). For instance, a brain MRI volume is drawn toward “brain MRI with Pituitary Tumor” and repelled from “Lung CT with Nodule.” During this process, both the text and 2D image encoders remain frozen.

\subsection{Contrastive Loss Function}
To effectively align 2D/3D image and text embeddings within a shared space, we utilize the Information Noise Contrastive Estimation (InfoNCE) loss \cite{infonce}. InfoNCE is one of the most common loss functions in contrastive learning—used in models such as CLIP, SimCLR, and MoCo, and widely adopted in recent cross-modality and contrastive learning research \cite{blankemeier2024merlinvisionlanguagefoundation, Yan_Tang_Zhang_Tang_2024, alayrac2022flamingovisuallanguagemodel, Ma_Liu_Han_Hu_Ju_2024}. It works by minimizing the distance between semantically similar image-text pairs while maximizing the distance between dissimilar ones. For example, when using 3D volumetric image embeddings, we calculate the cosine similarity between image-text pairs as part of this alignment process.

\[
\text{logits}_{ij} = \frac{\mathbf{V}_i \cdot \mathbf{T}_j}{\tau}
\]
where \(\mathbf{V}_i\) and \(\mathbf{T}_j\) are embeddings of the \(i\)th image and \(j\)th text, and \(\tau\) is a temperature parameter. In a batch of \(N\) pairs, a similarity matrix is formed with matching pairs along the diagonal and mismatched pairs elsewhere. The InfoNCE loss is defined as:
\[
\mathcal{L}_{i,j} = - \log \frac{\exp(\text{sim}(\mathbf{V}_i, \mathbf{T}_j) / \tau)}{\sum_{k=1}^{N} \exp(\text{sim}(\mathbf{V}_i, \mathbf{T}_k) / \tau)}
\]
where \(\text{sim}(\mathbf{V}_i, \mathbf{T}_j)\) denotes cosine similarity. The same loss function is used during 2D image encoder pre-training.

\subsection{Implementation Details}

We load pre-trained weights from the CLIP model (clip-vit-large-patch14) using the Hugging Face Transformers library. During RadCLIP pre-training, we employ a cosine annealing learning rate scheduler starting at 1e-4, save checkpoints at every epoch, and apply early stopping based on validation loss. Hyperparameters—including training epochs, learning rate, batch size, and attention head count—were empirically tuned \cite{clip}. To enhance robustness, dropout (rate 0.5) and L2 regularization were applied.

Training and evaluation were conducted on a system with 4 NVIDIA A6000 GPUs using PyTorch (v1.9) and Hugging Face Transformers (v4.12). The model weights, training and
evaluation code, and comprehensive documentation are now available on Hugging Face and
GitHub: https://github.com/luzhixiu/RadCLIP , ensuring reproducibility and enabling further research.
\section{Experiments}

\begin{figure*}[!t]
\centering
\includegraphics[width=\textwidth]{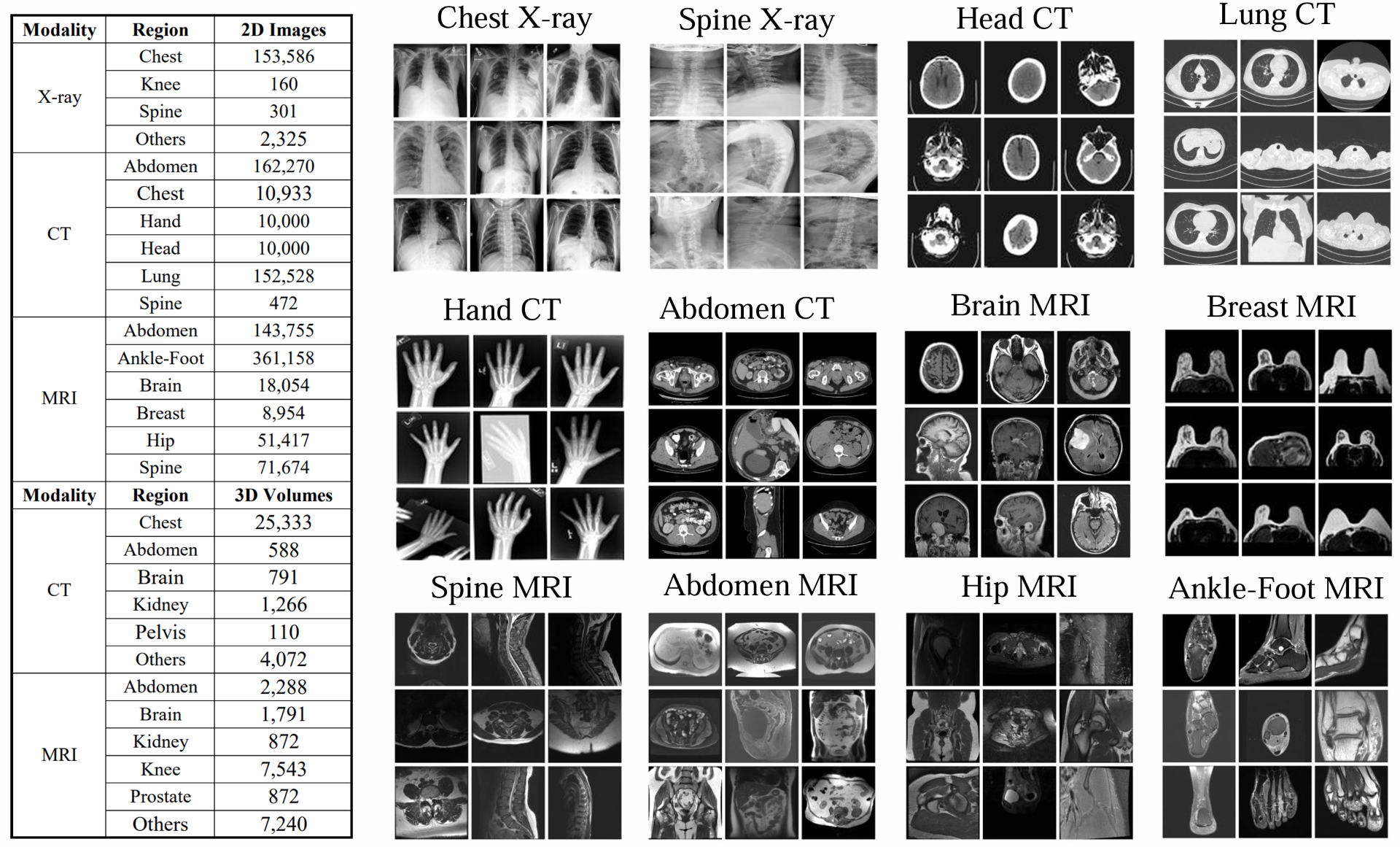}
\caption{Overview of the RadCLIP Datasets. This figure presents our comprehensive dataset, which includes 1,157,587 2D radiologic image–text pairs and 52,766 3D image–text pairs from 14 public sources. Representative samples illustrate the diversity in imaging modalities and anatomical regions used for training and evaluation.}
\label{fig_rad_img}
\end{figure*}

In this section, we describe the experimental configurations used to evaluate RadCLIP. We detail the datasets for training and evaluation, outline our evaluation strategy (including downstream tasks and metrics), and present our results.

\subsection{Dataset Curation}

To enable RadCLIP to learn from diverse radiologic images, we curated a large dataset from publicly available collections for pre-training. This training dataset comprises 1,157,587 2D image–text pairs (X-ray, CT, and MRI) and 52,766 3D image–text pairs (CT and MRI). It covers various anatomical regions and 124 distinct diseases and conditions, with “normal” being the most frequent label. The dataset was assembled from 14 public collections. Figure~\ref{fig_rad_img} shows the sample sizes and representative images from different modalities and body parts. We gratefully acknowledge the studies that made these datasets publicly available.

Additionally, we compiled an evaluation dataset from four public sources. These images were not part of the training set, serving as unseen external data for assessing generalization. The individual datasets are listed below.

\begin{itemize}
    \item \textbf{RadCLIP training dataset:}
    \begin{itemize}
        \item Cancer Moonshot Biobank - Colorectal Cancer Collection (CMB-CRC) \cite{cmbcrc2023}
        \item Cancer Moonshot Biobank - Lung Cancer Collection (CMB-LCA) \cite{cmblca2023}
        \item MOS-MED \cite{morozov2020mosmeddatachestctscans} 
        \item Duke-Abdomen \cite{wang2023dukespleendataset}
        \item ISPY1 \cite{ispy12023}
        \item NYU fastMRI \cite{fastMRI,fastMRIarxiv} 
        \item Open Neuro: Flanker Task \cite{flankertask}
        \item PI-CAI \cite{Saha2024}
        \item Prostate-MRI-US-Biopsy \cite{Natarajan2020}
        \item qDESS Knee MRI \cite{qDESS}
        \item RSNA Pneumonia \cite{RSNAPneumonia}
        \item RadImagenet \cite{radimagenet}
        \item Unifesp \cite{UNIFESP}
        \item CPTAC-PDA \cite{cptacpda2023}
        \item MedMNIST \cite{medmnist}
    \end{itemize}
\end{itemize}

\begin{itemize}
\item \textbf{RadCLIP evaluation dataset:}
    \begin{itemize}
        \item ChestXpert \cite{irvin2019chexpert}
        \item Crystal Clean Brain Tumor \cite{CrystalClean}
        \item IXI Brain \cite{IXIDataset}
        \item COVID-CT-MD \cite{afshar2021}
    \end{itemize}
\end{itemize}

All images were resized to $224 \times 224$ pixels. For 3D volumetric images, size normalization was performed on each acquisition plane (axial, coronal, and sagittal), and intensities were standardized using z-score normalization.

For each 2D/3D image or volume, we extracted descriptive text from associated documents and labels. These descriptions follow the pattern [body region – imaging modality – disease/medical condition (if applicable)], e.g., [Abdomen CT with prostate lesion] or [Brain MRI with Pituitary Tumor]. Not all texts include disease information. After curation, we tokenized all descriptions using CLIP’s default tokenizer.

\subsection{Evaluation Strategy}

After pretraining, we evaluated RadCLIP’s performance on downstream tasks, focusing on image classification and image–text matching using 2D and 3D radiologic images.

\begin{figure}[h]
\centering
\includegraphics[width=3in]{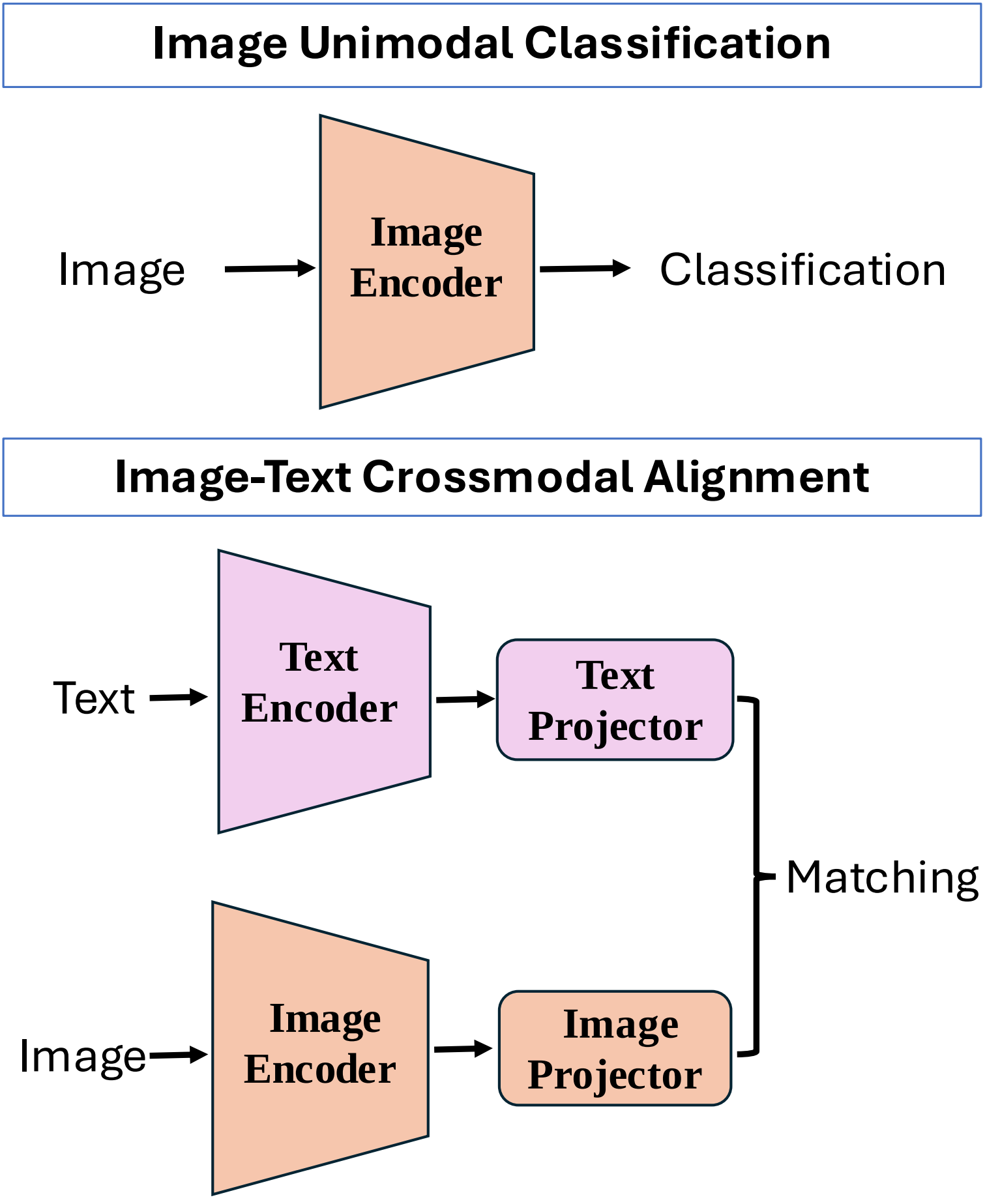}
\caption{Downstream Tasks Using RadCLIP. Top panels (Image Unimodal Classification) demonstrate the linear probing approach for image classification, where a single-layer classifier is trained on features extracted by RadCLIP. Bottom panels (Image–Text Crossmodal Alignment) illustrate the image–text matching setup using cosine similarity to align image embeddings with their corresponding textual descriptions.}
\label{fig_downstream}
\end{figure}

For image classification, we employed a linear probing strategy. RadCLIP’s 2D image encoder (with a slice pooling adapter) was used as a feature extractor, and a single-layer linear classifier was trained on the extracted features (see Figure~\ref{fig_downstream} A-B). This approach assesses the model’s radiologic image representations without fine-tuning the entire network. Experiments were conducted using five-fold cross-validation on the evaluation datasets, with splits of 70\% training, 10\% validation, and 20\% testing. Only the linear classifier was updated during training, and performance metrics were averaged across folds to assess robustness and generalizability.

For image–text matching, the model aligns image embeddings with corresponding text from several candidates. We computed the cosine similarity between the embeddings of 2D/3D images and all text candidates (see Figure~\ref{fig_downstream} C-D), and evaluated performance using top-1 precision.

We compared RadCLIP with several state-of-the-art models in medical image analysis, including ResNet50, Vision Transformer (ViT), Swin Transformer (SwinT), SimCLR, MoCo V2, and MedViT. We also evaluated vision–language models such as CLIP, CoCa, and PMC-CLIP.

\subsection{Results}

\subsubsection{Unimodal Image Classification Performance}

\begingroup
\let\clearpage\relax


\begin{table*}[!t]
\centering
\fontsize{7.5}{6}\selectfont 
\setlength{\tabcolsep}{0pt} 

\caption{Unimodal classification performance of RadCLIP compared to existing methods across multiple datasets.}
\label{table:classification} 

\begin{tblr}{
  cells = {c},
  cell{1}{3} = {c=2}{},
  cell{1}{5} = {c=2}{},
  cell{1}{7} = {c=2}{},
  cell{1}{9} = {c=2}{},
  vline{4,6,8} = {1}{},
  vline{2-3,5,7,9} = {2-12}{},
  hline{1-3,13} = {-}{},
}
 &  & {\textbf{ChestXpert~}\\\textbf{(5 Classes, 2D)}} &  & {\textbf{Crystal Clean~}\\\textbf{(4 Classes, 2D)}} &  & {\textbf{IXI Brain~}\\\textbf{(2 Classes, 3D)}} &  & {\textbf{COVID-CT-MD}\\\textbf{(3 Classes, 3D)}} & \\
\textbf{Model~Name} & \textbf{VLP} & \textbf{Acc (\%)} & \textbf{F1 (\%)} & \textbf{Acc (\%)} & \textbf{F1 (\%)} & \textbf{Acc (\%)} & \textbf{F1 (\%)} & \textbf{Acc (\%)} & \textbf{F1 (\%)}\\
ResNet50 \cite{ResNet} & N & 41.98 ± 1.16 & 41.74 ± 4.20 & 65.67 ± 2.71 & 57.73 ± 23.28 & 92.65~± 1.23 & 91.65~± 1.23 & 58.93 ± 11.01 & 51.43 ± 13.34\\
ViT \cite{vit}& N & 45.02 ± 1.97 & 44.80 ± 5.51 & 72.67 ± 6.20 & 71.41 ± 13.14 & 94.15 ± 1.53 & 94.09 ± 0.83 & 62.07 ± 5.83 & 57.07 ± 11.45\\
SwinT \cite{swin} & N & 44.48 ± 1.27 & 44.27 ± 5.60 & 70.67 ± 4.78 & 69.38 ± 14.10 & 92.58 ± 0.44 & 92.57 ±~0.44 & 63.31~± 6.07 & 57.56 ± 9.24\\
SimCLR \cite{simpleCLR} & N & 45.59~± 2.13 & 44.51~± 5.05 & 70.22~± 6.12 & 69.84 ± 13.20 & 94.15 ± 1.53 & 94.09 ± 0.83 & 62.79 ± 7.24 & 56.79 ± 6.27\\
MoCo V2 \cite{moco} & N & 46.27~± 2.55 & 46.20~± 4.72 & 71.54~± 7.48 & 70.63 ± 15.30 & 93.21 ± 0.84 & 93.19 ± 1.71 & 61.32 ± 9.24 & 54.79 ± 10.24\\
MedViT \cite{medvit} & N & 47.42~± 1.33 & 46.95~± 4.93 & 72.59 ± 6.52 & 71.77 ± 13.99 & 94.76~± 1.67 & 94.26~±~0.67 & 61.95 ± 6.59 & 55.53 ± 12.64\\
CLIP \cite{clip}& Y & 41.44 ± 2.23 & 40.44 ± 9.54 & 81.00 ± 3.59 & 80.42 ± 8.11 & 94.21 ± 0.11 & 93.88 ± 0.19 & 63.93 ± 6.13 & 56.58 ± 11.45\\
CoCa \cite{coca}& Y & 42.51~± 1.85 & 41.53~~± 10.08 & 78.27 ± 5.22 & 79.33 ± 10.87 & \textbf{96.11~± 0.91} & \textbf{96.11~± 0.91} & 62.95 ± 4.93 & 55.09 ± 9.92\\
PubMedCLIP \cite{pmcclip}& Y & 48.60~± 1.64 & 46.63~± 5.53 & 81.35~± 3.79 & 79.63 ± 9.79 & 95.76~±~0.67 & 95.76~±~0.67 & 57.70 ± 6.51 & 53.32 ±~6.99\\
RadCLIP (ours) & Y & \textbf{51.46 ± 1.32} & \textbf{51.54 ± 4.15} & \textbf{86.00 ± 6.02} & \textbf{87.11 ± 9.80} & 95.58 ± 1.49 & 95.57 ± 1.50 & \textbf{67.87~}±~\textbf{2.66} & \textbf{65.39~}±~\textbf{3.29}
\end{tblr}
\end{table*}
\endgroup

We evaluated unimodal image classification on four external datasets: ChestXpert, Crystal Clean, IXI Brain, and COVID-CT-MD (see Table~\ref{table:classification}).

On ChestXpert, models classified five diseases (Pneumothorax, Pleural Effusion, Edema, Atelectasis, and Lung Lesion) from 2D CXR images using a 5,000-sample evaluation set \cite{gloria,medclip}. RadCLIP achieved the highest accuracy (51.46\%) and F1 score (51.54\%), with PMC-CLIP and MedViT ranking second in accuracy (48.60\%) and F1 score (46.95\%), respectively.

For the Crystal Clean dataset, which classifies four brain conditions (Normal, Pituitary Tumor, Meningioma, and Glioma) from 2D brain MRI images, RadCLIP recorded the best accuracy (86.00\%) and F1 score (87.11\%). PMC-CLIP achieved the second-best accuracy (81.35\%), while CLIP obtained the second-best F1 score (80.42\%). Notably, vision–language models generally outperformed pure vision models.

On the IXI Brain dataset, which distinguishes gender from 3D T1 MRI images, CoCa led with 96.11\% accuracy and F1 score, while RadCLIP was a close second with 95.58\% accuracy and 95.57\% F1.

For the COVID-CT-MD dataset, classifying three lung conditions (Normal, COVID, and Pneumonia) from 3D CT images, RadCLIP achieved the best accuracy (67.87\%) and F1 score (65.39\%).

Overall, RadCLIP outperformed or matched other foundation models across all evaluation datasets, demonstrating its ability to generate robust 2D/3D radiologic image representations.

\subsubsection{Cross-Modal Image–Text Matching}

\begingroup
\let\clearpage\relax
\begin{table}[!t]
\centering
\fontsize{7.5}{8}
\setlength{\tabcolsep}{3pt} 
\caption{Cross-Modal Image-Text Matching Performance: Top 1 Precision (\%) for different models across various datasets.}
\label{table:image_text_matching}

\begin{tabular}{c|c|c|c|c} \hline
\textbf{Models} & \textbf{ChestXpert~} & \textbf{\textbf{Crystal~}Clean~} & \textbf{IXI Brain~} & \textbf{COVID-CT-MD} \\ \hline
CLIP & 20.41 & 15.83 & 50.18 & 19.67 \\
CoCa & 20.32 & 18.04 & 49.46 & 30.82 \\
PubMedCLIP & 19.11 & 15.83 & 55.12 & 40.33 \\
RadCLIP & \textbf{23.90} & \textbf{27.22} & \textbf{57.07} & \textbf{51.15} \\ \hline
\end{tabular}
\end{table}

\endgroup

\begin{figure*}[!t]
\centering
\includegraphics[width=\textwidth]{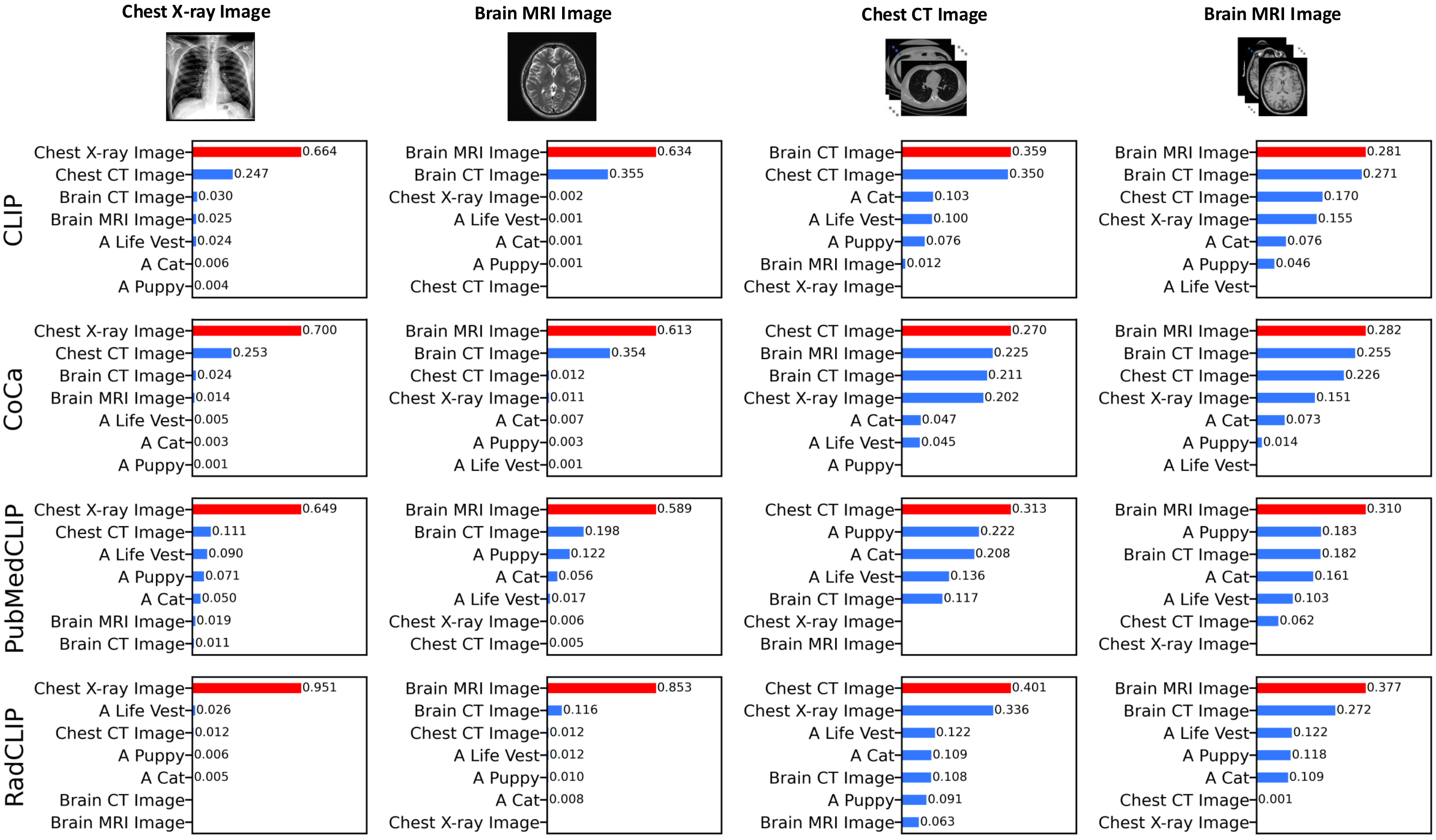}
\caption{Sample images from each benchmark dataset are paired with both correct modality labels (e.g., “Chest X-ray Image,” “Brain MRI Image,” “Chest CT Image”) and distractor labels (e.g., “A Puppy,” “A Cat,” “A Life Vest”). The accompanying bar charts show each model’s matching score for these text prompts. Higher scores indicate stronger alignment between the image and text.}
\label{vqa}
\end{figure*}

To assess RadCLIP’s cross-modal proficiency, we employed image–text matching as a downstream task. For instance, given an MRI image showing a brain glioma, a robust model should produce an image embedding closer to the text embedding for “brain glioma tumor” than to that for “normal brain.”

We evaluated this task using external datasets. For models with only a 2D image encoder, we applied global average pooling to adapt to 3D inputs. In this experiment, we compared RadCLIP with other vision–language models (CLIP, CoCa, and PMC-CLIP). Top-1 precision results are presented in Table~\ref{table:image_text_matching}: RadCLIP achieved 23.90\% on ChestXpert, 27.22\% on Crystal Clean, 57.07\% on IXI Brain, and 51.15\% on COVID-CT-MD, consistently outperforming its peers. CoCa often ranked second in precision.

These results demonstrate RadCLIP’s ability to effectively bridge visual and textual information in radiology, highlighting its potential for applications such as automated report generation and diagnostic assistance.

To further illustrate RadCLIP’s capacity for image–text matching, we conducted a simplified experiment focused on modality recognition. We selected the first image from each benchmark dataset (e.g., chest X-ray, brain MRI, chest CT) and paired them with correct labels (e.g., “Chest X-ray Image”) and distractor labels (e.g., “A Puppy,” “A Cat,” “A Life Vest”). As shown in Figure~\ref{vqa}, RadCLIP consistently identified the correct modality with higher confidence than other models, underscoring its robust alignment of visual and textual representations even in basic discrimination tasks.

\subsubsection{Ablation Study of RadCLIP Components}

\begingroup
\let\clearpage\relax

\setlength{\tabcolsep}{6pt} 

\begin{table*}[!t]
\centering
\fontsize{7.5}{12}\selectfont 
\definecolor{MineShaft}{rgb}{0.141,0.141,0.141}
\caption{Ablation Study of Different Pretraining Setup, Accuracy (\%) and F1 scores (\%) are included for classification performance, and Top 1 Precision (\%) is included for image-text matching}
\label{fig:ablation_results}

\begin{tabular}{c|cc|cc|c|c} \hline
\multirow{2}{*}{\textbf{\textbf{\textbf{\textbf{\textbf{\textbf{\textbf{\textbf{\textbf{\textbf{\textbf{\textbf{\textbf{\textbf{\textbf{\textbf{\textbf{\textbf{\textbf{\textbf{\textbf{\textbf{\textbf{\textbf{\textbf{\textbf{\textbf{\textbf{\textbf{\textbf{\textbf{\textbf{\textbf{\textbf{\textbf{\textbf{\textbf{\textbf{\textbf{\textbf{\textbf{\textbf{\textbf{\textbf{\textbf{\textbf{\textbf{\textbf{\textbf{\textbf{\textbf{\textbf{\textbf{\textbf{\textbf{\textbf{\textbf{\textbf{\textbf{\textbf{\textbf{\textbf{\textbf{\textbf{Pretraining Setup}}}}}}}}}}}}}}}}}}}}}}}}}}}}}}}}}}}}}}}}}}}}}}}}}}}}}}}}}}}}}}}}} & \multicolumn{2}{c|}{\textbf{\textbf{IXI Brain}}} & \multicolumn{2}{c|}{\textbf{\textbf{COVID-CT-MD}}} & \textbf{\textbf{\textbf{\textbf{\textbf{\textbf{\textbf{\textbf{IXI Brain}}}}}}}} & \textbf{\textbf{COVID-CT-MD}} \\ \cline{2-7}
 & \textbf{Acc (\%)} & \textbf{F1 (\%)} & \textbf{Acc (\%)} & \textbf{F1~(\%)} & \textbf{\textbf{P@1~(\%)}} & \textbf{\textbf{P@1~(\%)}} \\ \hline
CLIP + Global Average Polling & 94.21 & 93.88 & 63.93 & 56.58 & 50.18 & 19.67 \\ \hline
CLIP + Trained Slice Pooling Adapter & 94.89 & 94.89 & 64.58 & 58.01 & 55.30 & 23.43 \\ \hline
RadCLIP (Fine-Tuned 2D Image Encoder) +~Global Average Polling & 95.07 & 94.93 & 66.54 & 64.73 & 54.95 & 50.20 \\ \hline
\textcolor[rgb]{0.141,0.141,0.141}{RadCLIP (}Fine-Tuned 2D Image Encoder + Trained Slice Pooling Adapter\textcolor[rgb]{0.141,0.141,0.141}{)} & \textbf{95.58} & \textbf{95.57} & \textbf{67.87} & \textbf{65.39} & \textbf{57.07} & \textbf{51.15} \\ \hline
\end{tabular}

\end{table*}
\endgroup

We conducted an ablation study (see Table~\ref{fig:ablation_results}) to assess the contributions of RadCLIP’s components. First, we established a baseline using the original CLIP image and text encoders (with global average pooling for 3D images) without domain-specific fine-tuning. Next, we evaluated the impact of adding our slice pooling adapter to the vanilla image encoder. Then, we assessed the effect of fine-tuning by replacing the vanilla image encoder with a 2D encoder pre-trained on radiology-specific image–text pairs (still using global average pooling for 3D images).

Each modification resulted in modest gains compared to the original CLIP, suggesting that both the 2D image encoder and the slice pooling adapter play key roles in improving performance. When combined in the full RadCLIP setup—with a fine-tuned 2D encoder and a pre-trained slice pooling adapter—the model achieved the best results, highlighting the benefits of integrating both components.
\section{Conclusion}

The integration of RadCLIP into radiologic image analysis marks a significant advancement in medical imaging. Leveraging the VLP framework of CLIP, RadCLIP effectively bridges the gap between radiologic images and textual data. Its ability to align 2D/3D radiologic images with their corresponding text annotations not only enhances diagnostic accuracy but also streamlines clinical workflows through robust, interpretable image representations. Furthermore, our experiments demonstrate that RadCLIP can offer enhanced diagnostic support and improved radiologic image-text correlation, thereby providing a foundation for future research. This model could potentially be extended to integrate additional clinical data, develop specialized sub-models for various disease types, explore advanced multi-modal fusion techniques, and support applications such as radiologic report generation and radiologic image-text retrieval systems. Future investigations in these areas may help bridge the gap between computational insights and clinical decision-making, potentially contributing to more personalized and effective medical diagnostics.

However, RadCLIP does have limitations that merit further exploration. Our reliance on a diverse yet finite dataset may not capture the full spectrum of radiologic imaging variations encountered in clinical settings. In particular, the dataset currently omits certain imaging modalities, such as ultrasound and PET, which could affect the model’s generalizability in these areas. To address this, we plan to extend our dataset to include these modalities and are also exploring domain adaptation techniques to mitigate performance degradation when applying our model to new imaging types.

A significant design choice in our approach was the use of short, concise, and accurate textual labels. This strategy minimizes ambiguity and enhances consistency across the dataset, thereby bolstering label accuracy. However, this benefit comes with a trade-off: the limited length and detail of these labels may restrict the richness of the semantic associations the model can learn. In contrast, longer, free-style texts could capture subtle nuances and a wider range of diagnostic details, though they might also introduce variability and noise that could compromise model training and reliability.

Furthermore, public access to diverse medical reports is very limited, which constrains the availability of richly detailed textual data. Nonetheless, researchers have the opportunity to fine-tune RadCLIP using their own texts, potentially enhancing the model’s ability to learn deeper semantic associations and adapt to specific clinical contexts.

While our dataset spans a broad range of modalities and conditions, further validation with more extensive, real-world clinical data would be beneficial. Additionally, the 3D slice pooling mechanism, while innovative, introduces complexity in model training and interpretation, potentially necessitating additional computational resources and optimization techniques. The fixed textual encoder, although effective in preserving language understanding, may limit the model’s adaptability to evolving medical terminologies and nuanced diagnostic language over time. Our current training did not include less common imaging modalities such as ultrasound (US) and PET, limiting its application in these areas. However, our framework is designed for the future integration of these modalities. In addition, the model architecture supports subsequent fine-tuning, allowing researchers to incorporate their own domain-specific data to enhance performance and adapt the system to a wide array of clinical environments.

In summary, RadCLIP offers a promising approach to enhance radiologic image analysis through advanced vision-language pretraining techniques. The model's fine-tuned radiologic image encoder, along with its novel slice-wise attention mechanism, underscores its potential to improve diagnostic accuracy and efficiency in the medical imaging domain.RadCLIP excels in representing radiologic images and aligning these with textual descriptions, paving the way for integrated diagnostic tools. Future work will aim to expand the dataset, incorporate images from less common imaging types, enrich the textual data, refine the 3D pooling mechanism, and dynamically adapt the textual encoder to ensure RadCLIP continues to advance in medical imaging technology.



\end{document}